A Study of "Left Before Treatment Complete" Emergency Department Patients: An Optimized Explanatory Machine Learning Framework

Abdulaziz Ahmed[a][∗], Khalid Y.Aram[b], Salih Tutun[c]

[a]Department of Health Services Administration, School of Health Professions, The University of Alabama at Birmingham, Birmingham, Alabama, USA
[b]Department of Business Administration, Emporia State University, Emporia, Kansas, USA
[c]WashU Olin Business School, Washington University in St. Louis, St. Louis, MO, USA

Abstract

The issue of left before treatment complete (LBTC) patients is common in today's emergency departments (EDs). This issue represents a medico-legal risk and may cause a revenue loss. Thus, understanding the factors that cause patients to "leave before treatment is complete" is vital to mitigate and potentially eliminate these adverse effects. This paper proposes a framework for studying the factors that affect LBTC outcomes in EDs. The framework integrates machine learning, metaheuristic optimization, and model interpretation techniques. Metaheuristic optimization is used for hyperparameter optimization--one of the main challenges of machine learning model development. Three metaheuristic optimization algorithms are employed for

---

[∗] Corresponding author
Address: 1720 University Blvd, Birmingham, AL 35294
Phone: +1 (205) 598-3531
Email: aahmed2@uab.edu

optimizing the parameters of extreme gradient boosting (XGB), which are simulated annealing (SA), adaptive simulated annealing (ASA), and adaptive tabu simulated annealing (ATSA). The optimized XGB models are used to predict the LBTC outcomes for the patients under treatment in ED. The designed algorithms are trained and tested using four data groups resulting from the feature selection phase. The model with the best predictive performance is interpreted using SHaply Additive exPlanations (SHAP) method. The findings show that ATSA-XGB outperformed other mode configurations with an accuracy, area under the curve (AUC), sensitivity, specificity, and F1-score of 86.61%, 87.50%, 85.71%, 87.51%, and 86.60%, respectively. The degree and the direction of effects of each feature were determined and explained using the SHAP method.

# 1 Introduction

Emergency departments (EDs) are the main source of hospital admissions (Moore et al., 2017). When a patient visits an ED, different disposition modes are possible including admission as inpatient, discharged, expired, or transferred. Another important disposition status is when the patient chooses to leave before receiving medical care, during treatment, or against medical advice. Those patients are called left without being seen (LWBS), Left against medical advice (LAMA), and left subsequent to being seen (LSBS) (Fry et al., 2004; Gilligan et al., 2009; Hall and Jelinek, 2007; Smalley et al., 2021). The Fourth Emergency Department Benchmarking Alliance (EDBA) summit published a metric called left before treatment complete (LBTC), which is used throughout this study, to refer to LWBS, LAMA, and LSBS combined (Smalley et al., 2021).

Leaving ED before finishing treatment is associated with a higher risk of readmission and mortality (Mataloni et al., 2018; Tropea et al., 2012). Those patients are often severely ill and when they leave without receiving appropriate or timely medical care, their life might be at risk. In 2020, a 25-years-old patient visited the ED of Froedtert Hospital in Wauwatosa, Wisconsin. The patient had chest pain and after waiting for more than two hours, she left the hospital. The condition of the patient deteriorated and later on, she died on the same day (Linnane, 2020). In addition, many healthcare systems use the rate of LBTC patients as a quality measure of ED services (Asaro et al., 2005; Bolton et al., 2006; Mohsin et al., 2007a; Sun et al., 2007a). In 2012, the national U.S. quality forum defined ten indicators for healthcare service performance in EDs and considered the rate of LBTC as an indicator of timely and effective care (Smalley et al., 2021). Another important aspect of LBTC is the risk of hospitals not getting reimbursed for these visits, which may cause a substantial loss of revenue. Thus, the LBTC rate represents a challenge for healthcare systems and may indicate a decline in the quality of healthcare services provided. Therefore, investigating the factors that affect the rate of LBTC is crucial to improving the quality of care, efficiency, throughput, and revenue. This also helps healthcare practitioners develop strategies and programs to reduce the rate of LBTC patients.

The purpose of this study is to analyze the factors that affect the rate of LBTC patients and develop a machine learning model that predicts LBTC patients and discriminates them from patients who leave after being seen and treated (SAT). In the past few years, machine learning methods have been used in many healthcare applications related to performance improvement in the areas of, for instance, emergency care (Ahmed et al., 2022), organ transplant (Badrouchi et al., 2021), and mental health (Tutun et al., 2022). One of the most challenging tasks in developing machine learning models is that most algorithms have many parameters that need to be tuned to achieve desirable performance(Ahmed et al., 2022). In this paper, we propose a new approach that integrates adaptive tabu simulated annealing (ATSA) optimization algorithm with eXtreme Gradient Boosting (XGB). The new algorithm is called ATSA-XGB. The goal of ATSA is to optimize the parameters of XGB and then use the optimized model for predicting LBTC patients. Simulated annealing (SA) and adaptive simulated annealing (ASA) are used to optimize XGB and compared with ATSA. The proposed model can be used as a decision-support tool to help predict LBTC patient rates. The tool can be used by healthcare practitioners and hospitals to detect and develop strategies to reduce the LBTC rate, and in return, improve the quality of care and reduce the negative impact of LBTC in EDs. This study contributes to the theory and practice of healthcare analytics, the application of which creates value for healthcare operations. This study also contributes to the area of integrating machine learning and optimization by demonstrating how machine learning performance can be boosted using optimization techniques. More specifically, the theoretical and practical contributions of this study are as follows:

- An explanatory machine learning approach is proposed to study the factors that affect the rate of LBTC patients.
- A new approach is proposed to optimize the parameters of XGB based on ATSA.
- To the best of our knowledge, this is the first study that proposes ATSA for optimizing the hyperparameters of machine learning. Also, this is the first study that develops a comprehensive machine learning framework to predict LBTC status.
- An extensive number of experiments are conducted to evaluate the proposed ATSA-XGB algorithm (a total of 20 models are developed).

The remainder of this paper is organized as follows. Section 2 summarizes the relevant studies on LBTC and machine learning parameter optimization. Section 3 describes the proposed methodologies including data collection, preprocessing, and model development. The experimental results are presented in Section 4, and finally, the conclusion and future work are presented in Section 5.

## 2 Literature review

In this section, we first review recent studies on LBTC patients including. We then present studies that used different algorithms to optimize the performance of machine learning.

### 2.1 Relevant LBTC studies

Several studies proposed models to predict the disposition status of ED patients, most of which focused on the most common disposition statuses, which are admission and discharge (Afnan et al., 2021; Ahmed et al., 2022; Hong et al., 2018; Lee et al., 2020; Luo et al., 2019; Mowbray et al., 2020; Shafaf and Malek, 2019; Sterling et al., 2019). Numerous studies investigated the reasons and consequences of LBTC patients including LWBS, LAMA, and LAMA. Table 1 provides a summary of relevant LBTC studies. Most of the studies focused on either LWBS or LBTC in general, except for one study that compared LWBS and LAMA patients. Most of the studies performed univariate analysis to compare the LWBS or LBTC patients to patients with other disposition statutes. In all the univariate studies, the comparisons were based on patient demographics, medical conditions, and hospital characteristics. For example, (Baker et al., 1991) conducted a univariate analysis to investigate whether LWBS patients needed immediate treatment after leaving the ED. They used patient demographics such as age, sex, race, and other information including insurance status, chief complaint, and acuity level to perform the comparison. (Mohsin et al., 2007b) used univariate analysis to investigate if the LWBS patients receive alternative medical care after leaving. Although univariant analysis is simple and easy to understand, it is not comprehensive and does not consider the relationship between the different variables. Therefore, some studies used multivariate approaches to study LBTC patients

**Table 1:** Summary of the works that studied LWBS, and LBTC.

| Study | Sample size | Univariate Study | Multivariate study | Machine learning study | Predictors/characteristics | LWBS | LBTC or LAMA |
|---|---|---|---|---|---|---|---|
| Baker et al. (1991) | 186 + 211 | * | | | 8 | * | |
| Sheraton et al. (2020) | 32,680,232 | * | * | * | 13 | * | |
| Improta et al. (2021) | 83,739 | | | * | 5 | * | |
| Mohsin et al. (2007b) | 14 741 | * | | | 8 | * | |
| Sun et al., (2007b) | 810.6 M | * | | | 8 | * | |
| Carron et al. (2014) | 307,716 | * | | | 5 | * | * |
| Ding et al. (2007) | 3,624 | * | | | 10 | * | |
| Mataloni et al. (2018) | 835,440 | * | | | 12 | * | |
| Crilly et al. (2013) | 64292 | * | * | | 8 | * | |
| Tropea et al. (2012) | 239,305 | * | * | | 9 | * | |
| Tropea et al. (2012) | 14,937 | * | * | | 7 | * | |
| Goodacre and Webster (2005) | 71331 | | * | | 6 | * | |
| Pham et al. (2009) | 283,907 | * | * | | 6 | * | |
| Hitti et al. (2020) | 266 | * | * | | 8 | * | |
| Zubieta et al. (2017) | 42,750 | * | | | 4 | * | |
| Mohsin et al. (2005) | 4,356, 323 | * | * | | 11 | * | |
| Smalley et al. (2021) | 626,548 | * | | | 4 | | * |
| Arab et al. (2015) | 768 | * | | | 9 | | * |
| Natan et al. (2021) | 390 | * | * | | 5 | | * |
| Pages et al. (1998) | 2425 | * | * | | 7 | | * |
| Mitchell et al. (2021) | 561,823 | * | * | | 24 | | * |
| Adeyemi and Veri (2021) | 219564 | * | * | | 10 | | * |
| Jerrard and Chasm (2011) | 199 | * | | | 4 | | * |
| Myers et al. (2009) | 581,380 | * | | | 8 | | * |
| Myers et al. (2009) | 429 | * | | | 12 | | * |
| Villarreal et al. (2021) | 546,856 | * | | | 11 | | * |

| Our study | 478,212 | | * | 17 | * |

Most of the previous multivariate studies utilized logistic regression to study LBTC patients. Logistic regression was used to determine the factors that affect LBTC rates. For instance, (Mitchell et al., 2021) used a logistic regression model to identify the factors that affect the rate of LAMA or LBTC patients. Twenty factors were considered including patient diagnosis, race, gender, education level, and vital signs. Although logistic regression can capture the relationship between multiple inputs and a binary output, its performance declines as the number of features increases (Stoltzfus, 2011). In our study, we combine machine learning and optimization to develop a high-accuracy model to identify and explain the factors that affect the rate of LBTC patients. With machine learning, complex nonlinear relationships among the various variables can be captured (Chen and Asch, 2017). Machine learning methods overcome the problem of noisy data and can handle different types of variables including continuous, discrete, and text data (Badrouchi et al., 2021). Among the previous studies, only two studies used machine learning for investigating the LBTC problem. Sheraton et al. (2020) used the random forest to rank the features that affect the LBTC rate by importance. However, the accuracy of their model was not reported. (Improta et al., 2021) used random forest, naïve Bayes, Logistic regression, and support vector machines to predict LBTC patients. In their study, the random forest had the best performance with an accuracy of 86.52%. However, they did not interpret the features considered in their study. Our study provides both, a tool to predict LBTC patients and an interpretation of the features that affect LBTC disposition status.

## 2.2 Machine learning parameter optimization

Various methods can be used to optimize the hyperparameters of machine learning algorithms. Table 2 includes a summary of recent approaches proposed to optimize hyperparameters including Grid Search (GS), Random Search (RS), Bayesian optimization (BO), and metaheuristic-based approaches. A detailed review of machine learning hyperparameter optimization approaches introduced before 2020 was presented by (Yang and Shami, 2020). In GS, hyperparameters are mapped to a grid space and every possible combination of hyperparameters on the grid is used to fit and evaluate a machine learning model. A negative aspect of GS is that it has a high computational cost. The number of models trained and evaluated increases exponentially as the number of parameters increases. In RS, the search space is pre-specified by upper and lower bounds on hyperparameters values of a machine learning algorithm. Afterward, a machine learning model is fitted and evaluated using a random selection of hyperparameter values from within the bounds. A disadvantage of RS is the high variance of model performance (Andradóttir, 2015). BO has also been used for optimizing machine learning parameters (Guo et al., 2019; Snoek et al., 2015). The disadvantage of BO is that its performance deteriorates dramatically as the number of parameters increases.

**Metaheuristics such as genetic algorithm (GA) and simulated annealing (SA) have been utilized to optimize the hyperparameters of machine learning algorithms. An advantage of metaheuristics is their capability of dealing with non-convex, discrete, and non-smooth optimization problems. Some studies used population-based algorithms, while others used single-solution algorithms. Particle swarm optimization (PSO) and GA are the most popular population-based algorithms. For instance, PSO and GA have been used to optimize the regularization parameter C of support vector machines (SVM) (Chou et al., 2014; Pham and Triantaphyllou, 2011), the number of hidden layers in artificial neural networks (ANN) (Sarkar et al., 2019), and the hyperparameters of XGB (Chen et al., 2020). Other population-based algorithms including Artificial Bee Colony (ABC), Ant Lion Optimization (ALO), and Bat Algorithm (BA) were used to optimize a few hyperparameters of a convolutional neural network (CNN) (Gaspar et al., 2021). Also, moth flame optimization (MFO), gray wolf optimization (GWO), and whale optimization algorithm (WOA) are used to optimize the parameters of SVM (Zhou et al., 2021). Differential Flower Pollination (DFP) was used to optimize an SVM model (Hoang and Tran, 2019). A disadvantage of GA and PSO is the computational cost since they handle a population of solutions that carry different sets of hyperparameter values throughout the search (Lin et al., 2010). Single-solution algorithms have also been used for hyperparameter optimization. For instance, (Tsai et al., 2020) used SA to optimize the hyperparameters of deep neural networks (DNN) for**

predicting bus passengers. (Bereta, 2019) used Tabu Search (TS) to optimize the hyperparameters of Adaboost (ADAB). In addition to the aforementioned approaches, there are code libraries for hyperparameter optimization such as Spearmint, BayesOpt, Hyperopt, SMAC, and MOE, which include tools that are based on RS, BO, or a combination of both. More information about these libraries can be found in (Yang and Shami, 2020).

Table 2: Summary of studies that proposed methods for optimizing hyperparameters of machine learning.

| Author(s) | Machine learning Algorithm(s) | GS | RS | BO | TS | GA | PSO | ALO | BA | ABC | DFP | MFO | GWO | WOA | SA | ASA | ATSA |
|---|---|---|---|---|---|---|---|---|---|---|---|---|---|---|---|---|---|
| Bergstra et al. (2011) | DBNs | | * | | | | | | | | | | | | | | |
| Bergstra & Bengio (2012) | DBNs | | * | | | | | | | | | | | | | | |
| Li et al. (2017) | SVMs | * | * | | | | | | | | | | | | | | |
| Guo et al. (2019) | XGB | | | * | | | | | | | | | | | | | |
| Badrouchi et al. (2021) | LR, KNN, XGB, MLP | * | | | | | | | | | | | | | | | |
| Chou et al. (2014) | SVMs | | | | | * | | | | | | | | | | | |
| Pham & Triantaphyllou (2011) | SVM, ANN, DT | | | | | * | | | | | | | | | | | |
| Sarkar et al. (2019) | SVM, ANN | | | | | * | * | | | | | | | | | | |
| Chen et al. (2020) | XGB | | | | | * | | | | | | | | | | | |
| Tsai et al. (2020) | DNN | | | | | | | | | | | | | | * | | |
| Bereta (2019) | Adaboost | | | | * | | | | | | | | | | | | |
| Gaspar et al. (2021) | CNN | | | | | | * | * | * | * | | | | | | | |
| Bibaeva, (2018) | CNN | | | | | * | | | | | | | | | * | | |
| Hoang & Tran (2019) | SVM | | | | | | | | | | | * | | | | | |
| B. Guo et al. (2019) | DNN | | | | | * | * | | | | | | | | | | |
| Snoek et al. (2015) | NN | | | | * | | | | | | | | | | | | |
| Zhou et al., (2021) | SVM | | | | | | | | | | | | * | * | * | | |
| Our study | XGB | | | | | | | | | | | | | | * | * | * |

## 3 Methodology

This section presents the different parts of the research framework including data preprocessing, feature selection, model development, and model interpretation.

### 3.1 Research framework

Figure 1 illustrates the proposed framework. First, the data is obtained and preprocessed. The data includes all the events that describe an ED episode from arrival to discharge. In the preprocessing stage, the missing values are handled, and the features are encoded and scaled. In the feature selection stage, two feature selection methods are used: Random Forest (RF) and Decision Tree (DT). These two methods are implemented using two search algorithms: Sequential Forward Selection (SFS) and Sequential Backward Selection (SBS). Five data groups result from the feature selection step. Four groups are based on the feature selection methods, and one includes all the features (e.g., X_all). Each data group is used to train and test a predictive model during the model development, resulting in a total of 15 different models. ATSA, ASA, and SA are used to optimize the parameters of XGB. An 80% to 20% training-to-testing ratio is used. Finally, the model with the highest performance is interpreted using SHAP.

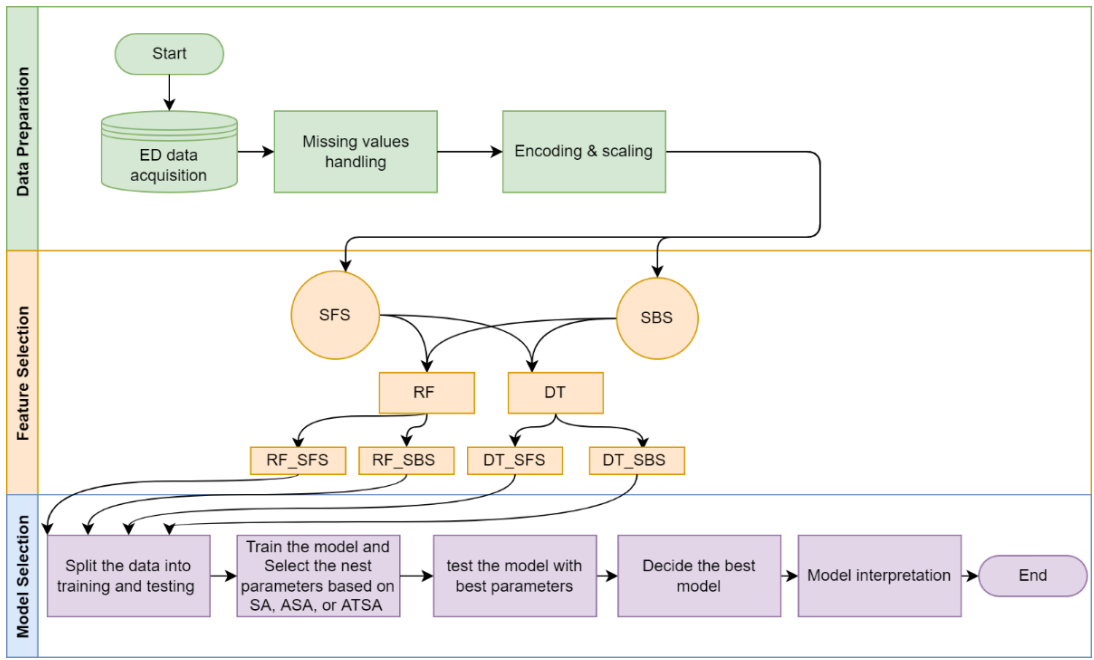

Figure 1: The proposed research framework.

## 3.2 Data collection and preprocessing

Retrospective patient-level data of ED visits between 2017 and 2019 is collected from a partner hospital located in the Midwest of the United States. The data includes more than 450k ED visits and more than 32 predictors. The following inclusion and exclusion criteria are applied to prepare the data for modeling:

- The timestamp features are excluded such as arrival and physician assessment times. For medical diagnoses, multiple features are included in the data, which are the International Classification of Diseases (ICD) codes, the description of the diagnoses, and the diagnosis group. To make the interpretation of the final model practical, only the diagnosis group is included.

- The arrival time feature is converted into three features: month (e.g., January), weekday (e.g., Saturday), and hours (e.g., 1-24). The minutes and the seconds are excluded.
- The demographic and vital signs features are included.

After implementing the above criteria, the final number of features comes to 18 including the output feature (See Table 3). Two more steps are taken to prepare the data for modeling: handling missing data and encoding categorial features. The percentages of missing data for all features are presented in Table 2. If all the visits with missing values are deleted, a significant amount of information will be lost. Therefore, K-Nearest Neighbor (KNN) is used for imputation. With a KNN imputer, missing values are replaced by the average of $K$ neighbors. After data imputation, patient sex and ethnicity are encoded using one-hot encoding, while the patient smoking feature is encoded using integer encoding. Integer encoding replaces each category in a feature with a unique integer number. The five classes of smoking status which are unknown, never, former, exposure, and current, are encoded as 0, 1, 2, 3, and 4, respectively. The output feature (disposition decision) includes two categories: SAT and LBTC patients. The SAT includes admitted and discharged patients, while LBTC includes LWBS and LAMA patients.

Table 3: Percentage of missing values of all features.

| Feature | Percentage of Missing Values |
|---|---|
| Respiratory Rate | 27.2% |
| O2 Saturation | 26.9% |
| Body Mass Index (BMI) | 25.7% |
| Systolic Blood Pressure | 25.7% |
| Diastolic Blood Pressure | 25.7% |
| Pulse Rate | 25.7% |
| Temperature in Fahrenheit | 25.7% |
| Emergency Severity Index (ESI) score | 26.6% |
| Patient Sex | 0.0% |
| Patient Age | 0.0% |
| Waiting time | 0.0% |
| Ed Department Location ID | 0.0% |
| ED Arrival Time hour | 0.0% |
| Patient Ethnicity | 0.0% |
| Patient Smoking Status | 0.0% |
| Month of year | 0.0% |
| Day of week | 0.0% |

## 3.3 Feature selection

Feature selection is an important step of model development. The goal of feature selection is to reduce the computational cost of model development and improve the generalization of models by excluding irrelevant features. In this paper, sequential feature selection methods (SFMs) are used for feature selection. The SFMs are greedy search algorithms used to select a subset of $k$ features from a dataset composed of $d$ features, where $k < d$. The features are added or removed sequentially based on model performance. Every time a feature is added or removed; the model performance is evaluated. The process continues until the model performance converges. Two sequential feature selection methods are used in this paper: 1) Sequential Forward Selection (SFS); 2) Sequential Backward Selection (SBS). The two methods are used along with two machine learning-based feature selection methods, which are DT and RF.

Consider a dataset with $d$ features ($Y = y_1, y_1, ..., y_d$) and $X_k$ is the subset of selected features, where $X_k = \{x_j \,|\, j = 1, 2, .., k; x_j \in Y\}$ and $K = 1, 2, ... d$. The pseudocode codes for SFS and SBS are shown in Algorithms 1 and 2, respectively. In SFS, a model starts with zero features such that $X_0 = \emptyset, k = 0$. In each iteration, a feature is added to the list of selected features $X_{k+1} = X_k + x^+$ and then the model is evaluated. If the model performance is improved, the feature is added and otherwise, it is excluded. The process continues until a termination criterion is met. In SBS, the model starts with all features $X_0 = Y, k = d$. In each iteration, a feature is removed from the list of selected features $(X_{k-1} = X_k - x^-)$ and the model performance is evaluated. The process continues until a termination criterion is met.

**Algorithm 1- SFS feature selection**
1. Input: $Y = (y_1, y_1, \ldots, y_d)$
2. Output: $X_k = \{x_j \,|\, j = 1, 2, \ldots, k; x_j \in Y\}$ and $K = 1, 2, \ldots d$
3. Start: $X_0 = \emptyset, k = 0$
4. While $k < d$
5.    Feature is added to $X_k$
6.    If $J(X_k + x^+) > J(X_k)$
7.       $(X_{k+1} = X_k + x^+)$
8.       $K = k + 1$
9.    Else:
10.    Go to step 2
11. End while
12. Return $X_k$

**Algorithm 2- SBS feature selection**
1. Input: $Y = (y_1, y_1, \ldots, y_d)$
2. Output: $X_k = \{x_j \,|\, j = 1, 2, \ldots, k; x_j \in Y\}$ and $K = 1, 2, \ldots d$
3. Start: $X_0 = Y, k = d$
4. While k <d
5.    Feature is added to $X_k$
6.    If $J(X_k - x^-) > J(X_k)$
7.       $(X_{k-1} = X_k - x^-)$
8.       $K = k - 1$
9.    Else:
10.    Go to step 2
12. End while
13. Return $X_k$

## 3.4 Extreme gradient boosting (XGB)

XGB, developed by (Chen and Guestrin, 2016), is a gradient boosting algorithm. It has competitive prediction performance and computational time compared to other well-known machine learning algorithms (Chen et al., 2020).

XGB obtains a strong learner ( i.e., a tree), based on the sum of predictions of multiple weak learners. This process can be summarized using the following equation:

$$\hat{y}_i = \varphi(x_i) = \sum_{k=1}^{K} f_k(x_i), \quad f_k \in F \qquad (1)$$

where $x_i$ is a vector of feature values, $y_i$ is the class label that corresponds to $x_i$, $K$ is a parameter that specifies the desired number of weak learners, and $f_k$ is the prediction score of the $k^{th}$ learner. Regularization is used to improve prediction performance as follows:

$$L(\Phi) = \sum_{i} l(\hat{y}_i, y_i) + \Omega(f_k) \qquad (2)$$

The part $\sum_i l(\hat{y}_i, y_i)$ is a loss function, which computes the difference between true ($y_i$) and predicted ($\hat{y}_i$) class labels. The term $\Omega(f_k)$ is a regularization function that penalizes the complexity of the model to prevent overfitting and can be formulated as follows:

$$\Omega(f_k) = \gamma T + \frac{1}{2} \lambda \sum_{j=1}^{T} w_j^2 \qquad (3)$$

where $T$ is the number of leaf nodes in the weak learner and the parameters $\gamma$ and $\lambda$ control the regularization. $\gamma$ is the coefficient of the number of leaves and the $\lambda$ is the coefficient of the *l*-2 norm of the weights of all the leaf nodes.

To further control overfitting and reduce computational complexity, XGB utilizes several randomization approaches such as column subsampling and random subsampling. One of the challenges of using XGB is its relatively large number of model parameters. These parameters include the learningRate, nEstimators, maxDepth, minChildWeight, $\gamma$, subSample, colSampleByTree. Optimizing these parameters is critical to achieving competitive prediction performance (Chen et al., 2020). To further improve prediction performance, this study attempts to fine-tune the parameters of XGB using ATSA, ASA, and SA.

### 3.5 Parameter optimization

Different machine learning methods have different hyperparameters and some parameters can have infinite possible values. For instance, a Support Vector Machine model with a radial basis kernel has two real-value hyperparameters: the regularization parameter $C > 0$ and the shape parameter $\gamma \in [0,1]$ (Couellan and Wang, 2017). To overcome this challenge, the selection of hyperparameter values for a given machine learning model can be modeled as a mathematical optimization problem (Equations 4 and 5). Suppose that $S$ is a vector of the parameters to be optimized such that $S = \{s_1, s_2, \dots, s_n\}$, and $n$ is the number of parameters, the objective function of the optimization and constraints are as follows:

$$\operatorname*{Max}_{\mathbf{x}} f(S) \tag{4}$$

subject to:
$$\psi_i^L \leq s_i \leq \psi_i^U \quad \forall \ i = 1, 2, \dots, n \tag{5}$$

where $f(S)$ is a desired model performance metric to be maximized, (e.g., accuracy), and $\psi_i^L$ and $\psi_i^U$ are lower and upper bounds on the parameter $s_i$, respectively. The value $s_i$ can be real, integer, or binary following the type of hyperparameters for the machine learning model being optimized. In this paper, we describe a method to optimize hyperparameters of machine learning models using SA, ASA, and ATSA.

### 2.2.1 Canonical simulated annealing

SA, developed by Kirkpatrick et al. (1983), is commonly used for solving combinatorial optimization problems. SA is inspired by the annealing process of metals, which is a heat treatment method used to improve metals' properties. In the annealing process, a metal is subjected to physical and sometimes chemical changes in its properties. The optimal arrangement of metal particles is achieved through the annealing process, which depends on the cooling rate of the heated metal. SA mimics the real annealing process by making iterative movements controlled by a temperature parameter and a cooling schedule. Algorithm 3 shows how SA can be used to optimize the parameters of a machine learning model. Let $S = \{s_1, s_2, \ldots, s_n\}$ denote a certain solution obtained by SA. SA starts with an initial solution $S_0$ (e.g., parameters' values). The initial solution becomes the current solution ($S_{cur}$) and then a new solution ($S_{new}$) is generated by making a small change in one or a combination of variables in the current solution ($S_{cur}$). The performance of the newly generated solution $f(S_{new})$ is compared with the performance of the current solution ($f(S_{cur})$. If the new solution outperforms the initial solution, (i.e., $f(S_{new}) > f(S_{cur})$), the new solution becomes the current solution ($S_{cur} = S_{new}$). On the other hand, if the new solution does not result in better performance, the new solution is accepted as the new solution with a probability $c$, where $c$ is calculated based on the current annealing temperature $T_i$ (Equation 7). This mechanism promotes search diversity and prevents the search from getting trapped in local maxima. Traditional SA starts with a large temperature and decreases by a cooling rate $\alpha$.

$$c = \frac{1}{e^{\frac{f(S_{new}) - f(S_{curr})}{T_i}}} \tag{6}$$

**Algorithm 3:** Simulated annealing for machine learning hyperparameter optimization

**Input:** Initial temperature $T_0$, cooling rate $\alpha$, minimum temperature $T_{min}$
**Output:** $S_{best}$

1. Start
2. Generate initial solution $S_0$
3. $S_{best} \leftarrow S_0$
4. $S_{cur} \leftarrow S_0$
5. While $T_i > T_{min}$:
6.     Generate a neighboring solution, $S_{new}$
7.     Calculate $f(S_{new})$
8.     **if** $f(S_{new}) > f(S_{cur})$
9.         $S_{cur} \leftarrow S_{new}$
10.     **else if** $Uniform(0,1) < c$
11.         $S_{cur} \leftarrow S_{new}$
12.     **if** $f(S_{cur}) > f(S_{best})$
13.         $S_{best} \leftarrow S_{cur}$
14.     $T_{i+1} = \alpha T_i$
15. End While
16. Return $S_{best}$
17. End

### 2.2.2 Adaptive tabu-simulated annealing for machine learning

In traditional SA, as the temperature declines throughout the search, the algorithm's ability to uphill climbing decays, resulting in a higher chance of falling into local optima. To overcome this challenge, ASA is modified through the addition of an adaptive cooling schedule. The adaptive cooling schedule adjusts the current search temperature based on the search trajectory. The adjustment can result in either cooling or possibly reheating. ASA controls the search temperature using the following function:

$$T_i = T_{min} + \lambda \cdot \ln(1 + r_i) \qquad (7)$$

where $T_{min}$ is the lowest value, the temperature is allowed to take, $\lambda$ is a coefficient to control the temperature incline rate, and $r_i$ is a counter for the number of consecutive uphill moves at iteration $i$. $r_i$ is calculated throughout the search as follows:

$$r_i = \begin{cases} r_{i-1} + 1 & if\ f(S_{new}) < f(S_{curr}) \\ r_{i-1} & if\ f(S_{new}) = f(S_{curr}) \\ 0 & if\ f(S_{new}) > f(S_{curr}) \end{cases} \qquad (8)$$

One of the main disadvantages of ASA is that it may keep visiting the same solution (i.e., cycling), which may lead to a trap in a local maxima. Therefore, a modification is introduced to ASA by adding a tabu list to keep track of recently visited solutions. ATSA takes the advantage of both, adaptive SA and Tabu Search (TS). That is, ATSA includes a mechanism for escaping local optima through adaptive temperature control and improves the search performance by keeping track of recently visited areas using a tabu list introduced by (Azizi and Zolfaghari, 2004). Algorithm 4 presents the ATSA. It starts by generating an initial solution of the model's hyperparameters, $S_0$, and setting it as the best solution so far. Afterward, the ATSA starts by generating neighboring solutions to the current solution at hand. Solution generation is done in a way that suits the type of hyperparameters of the machine learning model being optimized. For instance, uniform and gaussian distributions can be used to generate neighboring values for continuous hyperparameters. For integer hyperparameters, the random selection of integer increments, or decrements can be used. If the generated neighboring solution is in the tabu list, it is ignored, and a new solution is generated. All generated solutions are added to the current tabu list. If the new solution is better than the current solution at hand, it replaces it. If the current solution has better performance than the new solution, the new temperature is determined using equation 9 and the new solution is accepted with a probability calculated using equation7 solution. ATSA yields the best solution found after terminating the search when the temperature reaches below the preset minimum temperature. Azizi and Zolfaghari (2004) applied ATSA for job shop scheduling. In this study, we adapted the original ATSA and used it for hyperparameter optimization of machine learning algorithms. The adapted ATSA procedure is listed under Algorithm 4.

| | **Algorithm 4:** Adaptive Tabu Simulated Annealing |
|---|---|
| | **Input:** Initial temperature $T_0$, cooling rate $\alpha$, minimum temperature $T_{min}$, Tabu list size $l$ |
| | **Output:** $S_{best}$ |
| 1 | Start |
| 2 | Generate initial solution $S_0$ |
| 3 | $S_{best} \leftarrow S_0$ |
| 4 | $S_{cur} \leftarrow S_0$ |
| 5 | Initiate tabu list: $\Omega = \{S_{curr}\}$ |
| 6 | While $T_i > T_{min}$: |
| 7 |    Generate neighborhood solution ($S_{new}$) |
| 8 |    **if** $S_{new} \in \Omega$ |
| 9 |       Go to 5 |
| 10 |    **else** Add $S_{new}$ to the top of the tabu list |
| 11 |    **if** $|\Omega| > l$ |
| 12 |       remove the solution at the bottom of the tabu list |
| 13 |    Calculate $f(S_{new})$ |
| 14 |    **if** $f(S_{new}) > f(S_{cur})$ |
| 15 |       $S_{cur} \leftarrow S_{new}$ |
| 16 |       $r_i = 0$ |
| 17 |    **else if** $f(S_{new}) < f(S_{cur})$ |
| 18 |       $r_i = r_{i-1} + 1$ |
| 19 |       **if** $Uniform(0,1) < c$ |
| 20 |          $S_{cur} \leftarrow S_{new}$ |
| 21 |    **if** $f(S_{cur}) > f(S_{best})$ |
| 22 |       $S_{best} \leftarrow S_{cur}$ |
| 23 |    $T_{i+1} = \alpha T_i$ |
| 24 | End While |
| 25 | **Return** $S_{best}$ |
| 26 | End |

## 3.6 Explanatory machine learning

Machine learning methods often work as "black-box" such that they produce models that are not interpretable. Recently, multiple approaches have been proposed to improve the interpretability of machine learning models. Example approaches are LIME (Ribeiro et al., 2016), DeepLIFT (Shrikumar et al., 2017), and Layer-Wise Relevance Propagation (Bach et al., 2015). In these approaches, cooperative game theory equations are used to estimate the classic Shapley regression values (Shapley, 1988). Suppose that a feature value of an instance is considered as a "player", while the prediction of the instance is considered as the "payout" in a game. The way that the payout is divided fairly among features is determined through Shapely values (Molnar, 2020).

Another approach used for "black box" model interpretation is called *SHaply Additive exPlanations* (SHAP), which was proposed by (Lundberg and Lee, 2017). In SHAP, the classic Shapley values approach is combined with other agnostic techniques such as LIME and DeepLIFT. Each feature is assigned an importance score, which represents the change in the expected model prediction, given the value of that feature. In SHAP, the importance score for instance $i$ of a feature $X$ (i.e., $X(i)$) is calculated based on its effect on model prediction. First, all features excluding $X$ are considered and then the deviation from the expected prediction is calculated when adding the instance $X(i)$ to all the features (Lundberg and Lee, 2017). Therefore, with SHAP, each score comes from an aggregated set of marginal contributions for each instance, which leads to the prediction of that instance. In this study, each patient's admission disposition (e.g., SAT vs. LBTC) can be interpreted by aggregating the marginal contributions of features and calculating the importance of each feature by averaging the marginal contribution for all instances. In this paper, SHAP is used to interpret the best model to determine the factors that affect the disposition status.

## 4    Experimental results

This section presents the experimental results of both, feature selection and prediction. The importance and interpretation of the features of the best model are explained in this section as well.

### 4.1    Experimental settings of model optimization

Eight XGB parameters are optimized in this study (See Table 4). Uniform distribution is used to generate the initial values of the parameters, then normal distribution is used to update the parameters during the optimization phase. The mean and standard deviation of the normal distribution are 0 and 1, respectively. The default values are used for the other parameters. The parameters are optimized using SA, ASA, and ATSA, and the results are compared. Table 5 shows the parameters of SA, ASA, and ATSA. Some parameters are specific to an algorithm, while others are used in multiple algorithms. For example, the initial annealing temperature is specific to SA, while the same $T_{min}$ is used for ASA and ATSA. The number of iterations is set the same for the three algorithms (e.g., SA, ASA, ATSA).

Table 4: Parameter ranges for XGB algorithms.

| Parameter | Possible range | Experimental range | Type |
|---|---|---|---|
| Number of estimators | $[1, \infty)$ | $[1, 50]$ | Integer |
| Maximum depth | $[1, \infty)$ | $[1, 50]$ | Integer |
| Maximum delta step | $[1, \infty)$ | $[1, 50]$ | Integer |
| Number of parallel trees | $[1, \infty)$ | $[1, 50]$ | Integer |
| Learning rate | $(0,1]$ | $(0,1]$ | Float |
| L1 regularization | $(0,1]$ | $(0,1]$ | Float |
| L2 regularization | $(0,1]$ | $(0,1]$ | Float |
| Gamma | $(0, \infty]$ | $(0,50]$ | Float |

Table 5: SA and ASA parameters.

| Metaheuristic | Parameter | Value |
|---|---|---|
| SA | Initial temperature | 1000 |
| | temperature decrease ($\alpha$) | 0.1 |
| ASA, ATSA | $T_{min}$ | 2 |
| | Rate of temperature increase ($\beta$) | 2 |
| | Tabu length | 20 |
| SA, ASA, ATSA | # of iterations | 1000 |
| | Number of moves | 8 |
| | Initial solution generation | Uniform distribution |
| | Neighborhood search | Normal distribution |

## 4.2  Feature selection results

Table 6 shows the feature selection results. The total number of features increased from ?? to 33 due to the application of one-hot encoding. The selected features per each feature selection method are marked by (√). The last column of Table 6 shows the number of times each feature was selected by each of the four feature selection methods. Table 6 provides information about the importance of the selected features. Generally, the more often a feature is selected, the more important the feature is. For example, the patient smoking status feature is selected by the four methods, which implies that this feature is important. On the other hand, the day of the week is not selected by any method, which implies that this feature is not important with respect to the LBTC outcome. The last row in Table 5 shows the number of features selected by each feature selection method. For example, ten features were selected by the DT_SFS method.

Table 6: Feature Selection results.

| Feature | DT_SFS | DT_SBS | RF_SFS | RF_SBS | Total |
|---|---|---|---|---|---|
| Patient_Smoking_Status | √ | √ | √ | √ | 4 |
| Diastolic_Blood_Pressure | √ | √ | √ | √ | 4 |
| Patient_Ethnicity_Unavailable | √ | √ | √ | √ | 4 |
| Ed_Department_Location_Id | √ |  | √ | √ | 3 |
| Patient_Sex | √ |  | √ | √ | 3 |
| O2_Saturation |  | √ | √ | √ | 3 |
| Hour of a day | √ | √ | √ |  | 3 |
| Patient_Ethnicity_Hispanic_or_Latino | √ |  | √ | √ | 3 |
| Pulse_Rate |  | √ |  | √ | 2 |
| Temperature_In_Fahrenheit |  | √ |  | √ | 2 |
| Respiratory_Rate | √ | √ |  |  | 2 |
| ESI_Score | √ |  | √ |  | 2 |
| Patient_Ethnicity_Not Hispanic or Latino | √ |  | √ |  | 2 |
| BMI |  |  |  | √ | 1 |
| Month of a year |  | √ |  |  | 1 |
| Waiting_time for physician |  | √ |  |  | 1 |
| Age_Years |  |  |  |  | 0 |
| Systolic_Blood_Pressure |  |  |  |  | 0 |
| Day of a week |  |  |  |  | 0 |
| Total | 10 | 10 | 10 | 10 |  |

### 4.3 Optimization results

SA, ASA, and ATSA are used to optimize the parameters of XGB. In addition to all the features (X_all), the data groups obtained from the feature selection step are used to train and test the proposed algorithms, yielding a total of 15 models (5 data groups × 3 algorithms). AUC is used as the main performance measure during the optimization stage. Figure 2 shows the optimization convergence of ATSA-XGB, respectively. Most of the models converged after 800 iterations. The AUC of the converged models ranged between 68% and 86%. SA-XGB and ASA-XG converged in a similar manner. Tables 7 – 9 present the optimal parameters for SA-XGB, ASA-XGB, and ATSA-XGB, respectively. Regardless of the data group used, ATSA-XGB outperformed SA-XGB, and ASA-XGB, while ASA-XGB outperformed SA-XGB. The tabu list and the adaptive update of the annealing temperature are the main reasons for the enhanced performance of ATSA and its superiority compared to ASA and SA. The tabu list in the ATSA helps in avoiding the evaluation of recently visited solutions, (i.e., parameter sets), and exploring new solutions. The adaptive temperature update helps in avoiding local maxima traps as the annealing temperature is independent of the iteration number. The ATSA-XGB that is trained and tested using the RF_SBS data group resulted in the best model with an AUC of 86.61%. Further, it can be noticed that the XGB structure that resulted in the highest AUC, which is based on ATSA, has a shallow depth, but has the largest number of parallel trees among all the proposed models. SA-XGB and the data group produced by DT_SBS resulted in the highest AUC (81.62%), while in ASA-XGB, the groups DT_SFS, DT_SBS, and RF_SFS resulted in the highest AUCs (83.4%).

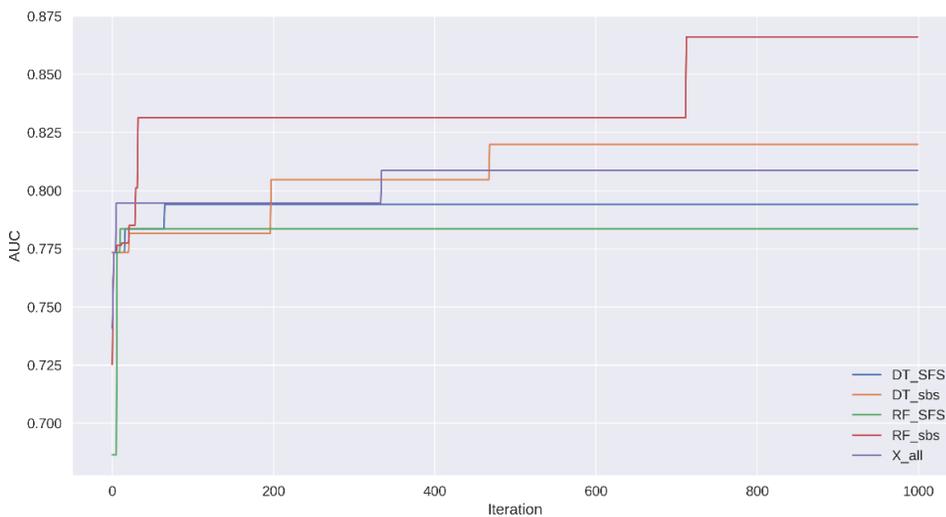

**Figure 2: ATSA-XGB convergence of all data groups.**

**Table 7: Optimal parameters for SA-XGB.**

| Parameter | DT_SFS | DT_SBS | RF_SFS | RF_SBS | X_all |
|---|---|---|---|---|---|
| Number of estimators | 3 | 2 | 1 | 1 | 1 |
| Maximum depth | 3 | 1 | 4 | 1 | 6 |
| Maximum delta step | 2 | 2 | 2 | 1 | 3 |
| Number of parallel trees | 2 | 3 | 2 | 3 | 1 |
| Learning rate | 0.80 | 0.19 | 0.88 | 0.97 | 0.62 |
| L1 regularization | 0.01 | 0.11 | 0.42 | 0.94 | 0.81 |
| L2 regularization | 0.49 | 0.94 | 0.11 | 0.39 | 0.02 |
| Gamma | 43.95 | 30.98 | 33.99 | 28.37 | 38.87 |
| AUC | 81.62% | 76.84% | 77.90% | 70.15% | 66.28% |

**Table 8: Optimal parameters for ASA-XGB.**

| Parameter | DT_SFS | DT_SBS | RF_SFS | RF_SBS | X_all |
|---|---|---|---|---|---|
| Number of estimators | 7 | 10 | 1 | 5 | 1 |
| Maximum depth | 1 | 3 | 12 | 9 | 6 |
| Maximum delta step | 6 | 1 | 1 | 7 | 3 |
| Number of parallel trees | 1 | 3 | 6 | 13 | 1 |
| Learning rate | 0.32 | 1.00 | 0.83 | 0.88 | 0.02 |
| L1 regularization | 0.02 | 0.43 | 0.82 | 0.87 | 0.12 |
| L2 regularization | 0.01 | 1.00 | 0.01 | 0.50 | 0.18 |
| Gamma | 12.26 | 33.89 | 10.71 | 41.81 | 16.46 |
| AUC | 83.84% | 83.84% | 83.84% | 82.94% | 83.54% |

**Table 9: Optimal parameters for ATSA-XGB.**

| Parameter | DT_SFS | DT_SBS | RF_SFS | RF_SBS | X_all |
|---|---|---|---|---|---|
| Number of estimators | 1 | 1 | 1 | 1 | 1 |
| Maximum depth | 1 | 10 | 1 | 1 | 1 |
| Maximum delta step | 1 | 2 | 1 | 1 | 4 |
| Number of parallel trees | 1 | 1 | 8 | 9 | 4 |
| Learning rate | 0.57 | 0.68 | 0.47 | 0.73 | 0.59 |
| L1 regularization | 0.78 | 0.16 | 0.00 | 0.71 | 0.03 |
| L2 regularization | 0.93 | 0.29 | 0.72 | 0.18 | 0.60 |
| Gamma | 5.48 | 20.01 | 3.95 | 7.03 | 9.80 |
| AUC | 79.41% | 81.99% | 78.36% | 86.61% | 80.87% |

## 4.4 Prediction results

Table 10 shows the AUC, accuracy, sensitivity, specificity, and f1 for the proposed models. Each row in the table represents a model. For example, the first row is named DT_SFS_SA-XGB, which refers to the SA-XGB algorithm trained and tested using the data group selected by the DT_SFS feature selection method. The ATSA-XGB model that is trained using the RF_SBS data group resulted in the highest AUC, hence it is the best model. It also resulted in the highest f1 (86.6%). The accuracy, sensitivity, and specificity of the best model are 87.50%, 85.71%, and 87.51%, respectively. Even though the best model did not have the highest accuracy, sensitivity, and specificity, it is still the strongest model because it can predict both LBTC and regular patients using severely imbalanced data. Figure 3 shows the confusion matrix for the best model. 85.7% of the LBTCs were detected. Given that LBTC patients represent only 6% of the dataset, the proposed algorithm achieved satisfactory and robust performance.

Table 10: Performance measures for all the proposed models.

| Model | Performance measure | | | | |
| --- | --- | --- | --- | --- | --- |
| | Accuracy | AUC | Sensitivity | Specificity | f1 |
| DT_SFS_SA-XGB | 96.40% | 81.62% | 66.67% | 96.58% | 78.88% |
| DT_SBS_SA-XGB | 86.90% | 76.84% | 66.67% | 87.02% | 75.50% |
| RF_SFS_SA-XGB | 89.00% | 77.90% | 66.67% | 89.13% | 76.28% |
| RF_SBS_SA-XGB | 73.60% | 70.15% | 66.67% | 73.64% | 69.98% |
| X_all_SA-XGB | 65.90% | 66.28% | 66.67% | 65.90% | 66.28% |
| DT_SFS_ASA-XGB | 82.00% | 83.84% | 85.71% | 81.97% | 83.80% |
| DT_SBS_ASA-XGB | 82.00% | 83.84% | 85.71% | 81.97% | 83.80% |
| RF_SFS_ASA-XGB | 82.00% | 83.84% | 85.71% | 81.97% | 83.80% |
| RF_SBS_ASA-XGB | 80.20% | 82.94% | 85.71% | 80.16% | 82.84% |
| X_all_ASA-XGB | 81.40% | 83.54% | 85.71% | 81.37% | 83.49% |
| DT_SFS_ATSA-XGB | 73.20% | 79.41% | 85.71% | 73.11% | 78.91% |
| DT_SBS_ATSA-XGB | 92.40% | 81.99% | 71.43% | 92.55% | 80.63% |
| RF_SFS_ATSA-XGB | 71.10% | 78.36% | 85.71% | 71.00% | 77.66% |
| RF_SBS_ATSA-XGB | 87.50% | 86.61% | 85.71% | 87.51% | 86.60% |
| X_all_ATSA-XGB | 76.10% | 80.87% | 85.71% | 76.03% | 80.58% |

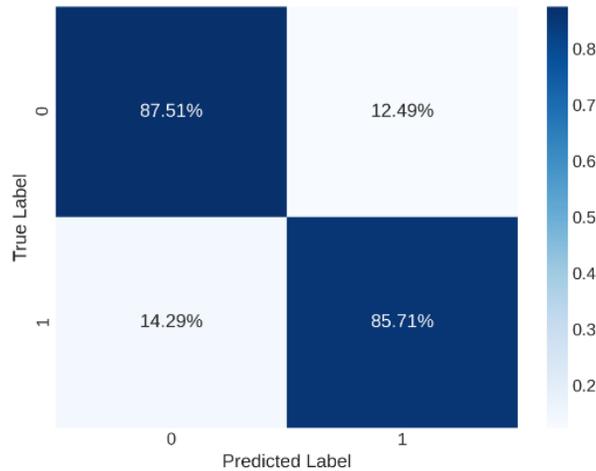

**Figure 3: Confusion matrix of the best model (SAT = 0, LBTC = 1).**

## 4.5 Model interpretation

To interpret the data features used in this study, we consider both the model that includes all the features and the best model found through selected features and hyperparameter optimization. Four models include all the features. The best model out of the four models has an AUC of 83.5% and was trained and tested using ASA-XGB (see Table 10). The model with optimized parameters is used along with the SHAP algorithm. The Python SHAP library is used to obtain feature scores and perform model interpretation. Instead of presenting the graphs generated by SHAP Library, we extracted the SHAP scores from the SHAP algorithm and then created Figure 4. The feature name is on the y-axis and the average SHAP score is on the x-axis. The features with positive SHAP scores are those that are positively correlated with the rate of LBTC, while the features with negative SHAP score are the features that are negatively correlated with the rate of LBTC.

Several observations can be made from Figure 4. Higher values of features such as age and pulse rate increase the chances of a patient *not* leaving the ED before treatment. In addition, some features increase the odds of LBTC. For example, the higher the waiting time for physicians, the higher the rate of LBTC patients. Further, the rate of LBTC patients increases during later times of the day. Looking at the severity of a patient's condition, the results show that higher ESI scores are associated with lower chances of patients leaving the ED before treatment. This means that patients with higher ESI scores (low acuity) are more likely to leave before treatment completion, which is consistent with what was found in the literature (Hitti et al., 2020; Rathlev et al., 2020)

The effects of a few features were surprising. For example, SHAP scores show that patients with low O2 saturation have higher chances of leaving before treatment is complete. Also, higher values of Diastolic and Systolic Blood Pressure and Body Temperature features are associated with higher chances of LBTC. This explains what is found in the literature patients leave with a risky condition and LBTC patients are at higher risk for readmission and mortality (Mataloni et al., 2018) (Tropea et al., 2012). Figure 5 shows the SHAP score for the features of the model with the highest AUC (e.g., best model). The features' impact on the rate of LBTC patients is consistent with shown in Figure 4.

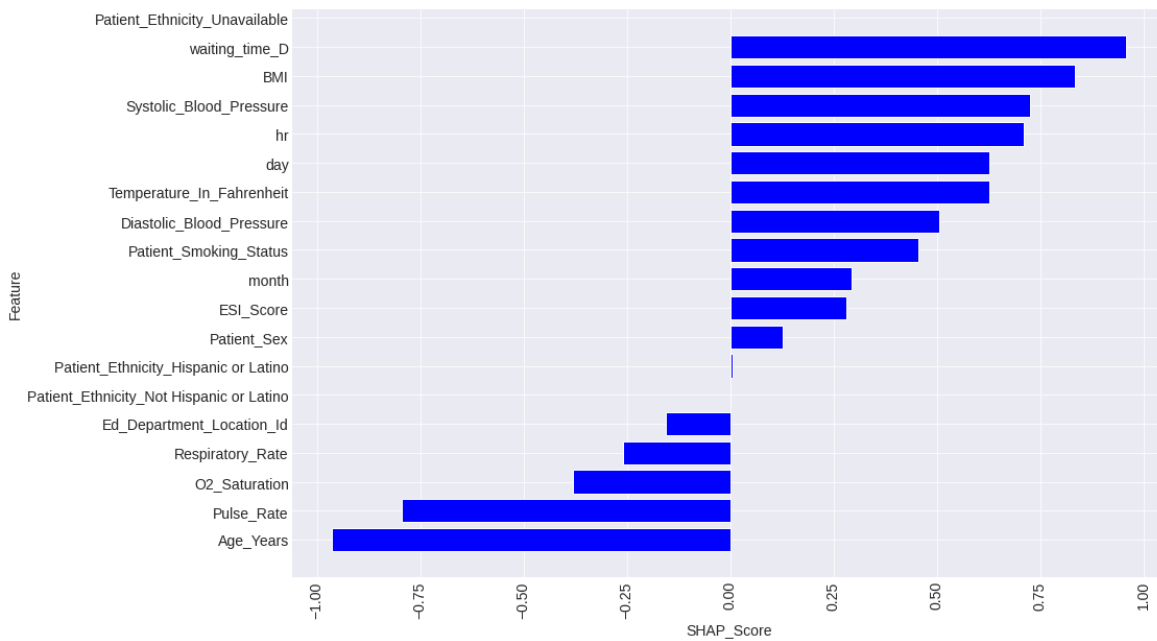

**Figure 4: SHAP scores for all features.**

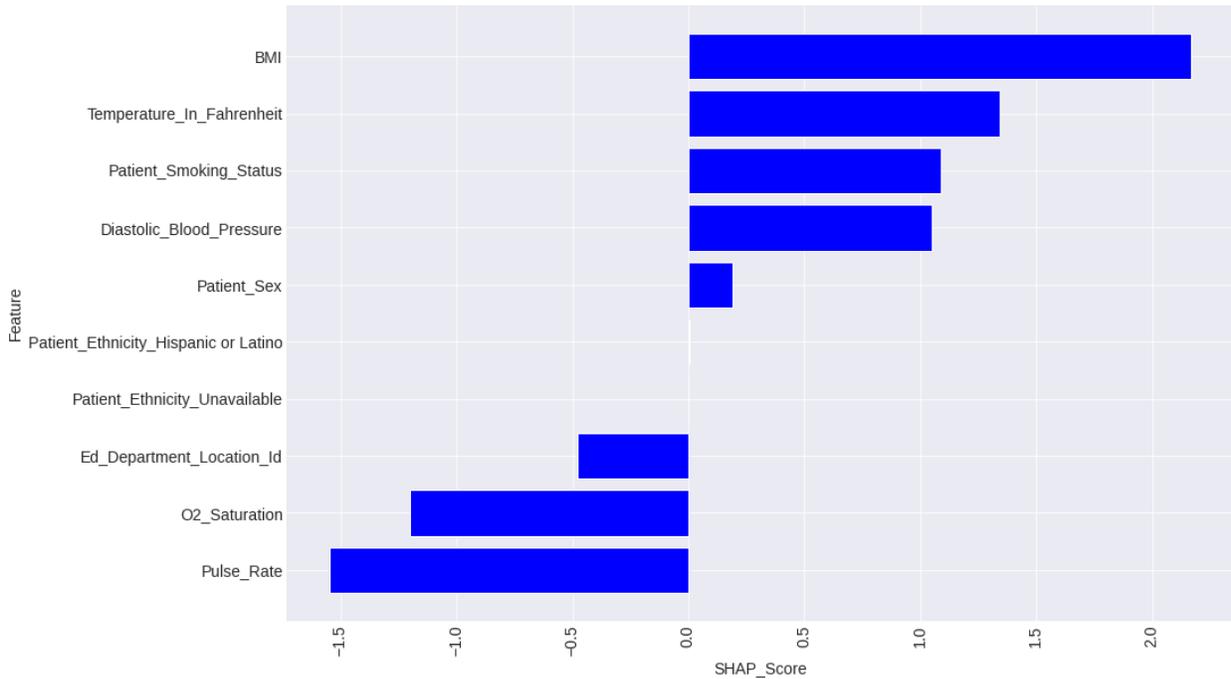

**Figure 5: SHAP scores for the features of the best model.**

## 5 Conclusions and future work

This paper presents an explanatory machine learning framework that can be used as a decision-support tool to help healthcare practitioners early identify LBTC patients in EDs. Early identification of LBTC patients allows for taking appropriate measures to prevent patients from leaving the ED before receiving necessary care. This would improve the quality of ED care and prevent adverse patient outcomes such as readmission and mortality. The proposed framework provides a theoretical and practical contribution to the area of healthcare analytics. The theoretical contribution comes from proposing the use of ATSA for optimizing XGB parameters. Although the proposed algorithm is used for optimizing XGB parameters, it can be used to optimize the parameters of any machine learning algorithm.

The practical contribution of this work is the application of machine learning to explain the effect of a patient's demographic information and vital signs on the rate of LBTC, which is an important quality and efficiency measure of ED processes. We developed various models based on different data groups that resulted from different feature selection procedures. The best model included only ten features and resulted in an AUC, accuracy, sensitivity, specificity, and f1 of 87.50%, 86.61%, 85.71%, 87.51%, and 86.60%, respectively. For future work, natural language processing can be used to analyze clinical notes, which may help in better understanding the causes of why patients leave the ED before treatment is complete.